\documentclass[10pt,journal,letterpaper,compsoc]{IEEEtran}
\def\IEEEcompilation{}

\ifCLASSOPTIONcompsoc
  \usepackage[nocompress]{cite}
\else
   \usepackage{cite}
\fi
\ifCLASSINFOpdf
   \usepackage[pdftex]{graphicx}
\else
   \usepackage[dvips]{graphicx}
\fi
\usepackage[cmex10]{amsmath}
\usepackage{amsfonts}
\usepackage{amssymb}
\usepackage{algorithm}
\usepackage{algorithmic}
\usepackage{array}
\ifCLASSOPTIONcompsoc
\usepackage[tight,normalsize,sf,SF]{subfigure}
\else
\usepackage[tight,footnotesize]{subfigure}
\fi
\usepackage{url}

\def\x{{\mathbf x}}

\def\l{{\ell}}
\def\1{{\mathbf 1}}

\def\p{{\mathbf p}}

\def\X{{\mathbf X}}
\def\Z{{\mathbf Z}}
\def\ZZ{{\mathcal Z}}

\def\betab{{\boldsymbol\beta}}

\def\alphab{{\boldsymbol\alpha}}

\def\y{{\mathbf y}}
\def\FFF{\text{F}}
\def\w{{\mathbf w}}
\def\D{{\mathbf D}}

\def\YY{{\mathcal Y}}

\def\W{{\mathbf W}}
\def\d{{\mathbf d}}
\def\E{{\mathbb E}}
\def\PPP{{\mathbb E}}
\def\EE{{\mathbf E}}
\def\O{{O}}

\def\s{{\mathbf s}}

\def\PPP{{\mathbb P}}

\def\d{{\mathbf d}}

\def\Real{{\mathbb R}}

\def\DD{{\mathcal D}}
\def\XX{{\mathcal X}}
\def\WW{{\mathcal W}}

\def\I{{\mathbf I}}

\def\argmin{\operatornamewithlimits{arg\,min}}

\def\trace{\operatorname{Tr}}

\def\sign{\operatorname{sign}}

\def\st{~~\text{s.t.}~~}
\def\defin{\stackrel{\vartriangle}{=}}

\newcommand{\citep}{\cite}
\newcommand{\citealp}{\cite}
\newcommand{\citet}{\cite}
\newtheorem{lemma}{Lemma} 
\newtheorem{proposition}{Proposition} 
\newcommand{\BlackBox}{\rule{1.5ex}{1.5ex}}  

\newcommand{\UpperSpace}{\vspace*{0.25cm}}
\newcommand{\LowerSpace}{\vspace*{0.25cm}}

\begin{document}
\title{Task-Driven Dictionary Learning}
\author{Julien~Mairal, Francis~Bach, and~Jean~Ponce 
   \IEEEcompsocitemizethanks{ \IEEEcompsocthanksitem 
      When this work was achieved, all authors were with INRIA - Willow Project-Team,
      Laboratoire d'Informatique de l'Ecole Normale Sup\'erieure.  (INRIA/ENS/CNRS
      UMR~8548), 23, avenue d'Italie, 75214 Paris. France.
      Jean Ponce is also with Ecole Normale Sup\'erieure, Paris.
      Francis Bach is now with INRIA - Sierra Project-Team. 
      Julien Mairal is now with the statistics department of the University of California, Berkeley.
      E-mail: \texttt{firstname.lastname@inria.fr} 
   }
}

\markboth{~}%
{Mairal \MakeLowercase{\textit{et al.}}: Task-Driven Dictionary Learning}


\IEEEcompsoctitleabstractindextext{%
   \begin{abstract}
Modeling data with linear combinations of a few elements from a \emph{learned}
dictionary has been the focus of much recent research in machine learning,
neuroscience and signal processing.  For signals such as natural images that
admit such sparse representations, it is now well established that these models
are well suited to \emph{restoration} tasks. In this context, learning the
dictionary amounts to solving a large-scale matrix factorization problem, which
can be done efficiently with classical optimization tools.  The same approach
has also been used for learning features from data for other purposes, e.g.,
image classification, but tuning the dictionary in a supervised way for these
tasks has proven to be more difficult.  In this paper, we present a general
formulation for supervised dictionary learning adapted to a wide variety of
tasks, and present an efficient algorithm for solving the corresponding
optimization problem.  Experiments on handwritten digit classification, 
digital art identification, nonlinear inverse image problems, and compressed
sensing demonstrate that our approach is effective in large-scale settings, and
is well suited to supervised and semi-supervised classification, as well as
regression tasks for data that admit sparse representations.
   \end{abstract}
   \begin{keywords}
       Basis pursuit, Lasso, dictionary learning, matrix factorization, 
       semi-supervised learning,
       compressed sensing.
   \end{keywords}
}

\maketitle

\IEEEdisplaynotcompsoctitleabstractindextext
\IEEEpeerreviewmaketitle

\section{Introduction}
\IEEEPARstart{T}he linear decomposition of data using a few elements from a \emph{learned}
dictionary instead of a predefined one---based on wavelets~\cite{mallat} for
example---has recently led to state-of-the-art results in numerous low-level
signal processing tasks such as image denoising~\cite{elad,mairal,mairal8},
audio processing~\cite{grosse,zibulevsky}, as
well as classification
tasks~\cite{raina,mairal3,mairal6,bradley,kavukcuoglu2,yang}.  Unlike
decompositions based on principal component analysis (PCA) and its variants, these
sparse models do not impose that the dictionary elements be orthogonal, allowing more
flexibility to adapt the representation to the data.

Consider a vector~$\x$ in~$\Real^m$. We say that it admits a sparse
approximation over a {\em dictionary}~$\D=[\d_1,\ldots,\d_p]$ in~$\Real^{m
\times p}$, when one can find a linear
combination of a ``few'' columns from~$\D$ that is ``close'' to the
vector~$\x$.  Experiments have shown that modeling signals with such sparse
decompositions ({\em sparse coding}) is very effective in many signal processing
applications~\cite{chen}.  For natural images, predefined dictionaries based on
various types of wavelets~\cite{mallat} have been used for this task.
Initially introduced by Olshausen and Field~\cite{field} for modeling the spatial receptive fields
of simple cells in the mammalian visual cortex, the idea of learning the
dictionary from data instead of using off-the-shelf bases has been shown to
significantly improve signal reconstruction~\cite{elad}. 

This classical data-driven approach to dictionary learning is well adapted to
reconstruction tasks, such as restoring a noisy signal. These dictionaries,
which are good at reconstructing clean signals, but bad at reconstructing
noise, have indeed led to state-of-the-art denoising
algorithms~\cite{elad,mairal,mairal8}. Unsupervised dictionary learning
has also been used for other purposes than pure signal reconstruction, such as
classification~\cite{grosse,raina,kavukcuoglu2,yang,wright2}, but recent works
have shown that better results can be obtained when the dictionary is tuned to
the specific task (and not just data) it is intended for. Duarte-Carvajalino and Sapiro~\cite{duarte} have for instance proposed
to learn dictionaries for compressed sensing, and in \cite{mairal3,mairal6,bradley}
dictionaries are learned for signal classification. In this
paper, we will refer to this type of approach as \emph{task-driven} dictionary
learning.

Whereas purely data-driven dictionary learning has been shown to be equivalent
to a large-scale matrix factorization problem that can be  effectively addressed
with several methods~\cite{field,engan,aharon,mairal7}, its
task-driven counterpart has proven to be much more difficult to optimize.
Presenting a general efficient framework for various task-driven dictionary
learning problems is the main topic of this paper. Even though it is different
from existing machine learning approaches, it shares similarities with
many of them.

For instance, Blei et al.~\citet{blei} have proposed to learn a latent topic model intended for
document classification. In a different context, Argyriou et al.~\cite{argyriou}
introduced a convex formulation for multi-task classification problems where an
orthogonal linear transform of input features is jointly learned with a
classifier.  Learning compact features has also been addressed in the
literature of neural networks, with restricted Boltzmann machines (RBM's) and
convolutional neural networks for
example~(see~\cite{lee4,ranzato,ranzato2,lecun,larochelle} and references therein).
Interestingly, the question of learning the data representation in an
unsupervised or supervised way has also been investigated for these approaches.
For instance, a supervised topic model is proposed in \citet{blei2}
and tuning latent data representations for minimizing a cost function is 
often achieved with \emph{backpropagation} in neural networks~\cite{lecun2}.

\subsection{Contributions}
This paper makes three main contributions:
\begin{itemize}
   \item It introduces a supervised formulation for learning dictionaries
      adapted to various tasks instead of dictionaries only adapted to data
      reconstruction.
   \item It shows that the resulting optimization problem is smooth under mild
      assumptions, and empirically that stochastic gradient descent addresses it
      efficiently.
   \item It shows that the proposed formulation is well adapted to
      semi-supervised learning, can exploit unlabeled data when
      they admit sparse representations, and leads to state-of-the-art results for
      various machine learning and signal processing problems.
\end{itemize}

\subsection{Notation}
Vectors are denoted by bold lower case letters and matrices by upper case ones.
We define for \mbox{$q\geq 1$}, the $\ell_q$-norm of a vector~$\x$ in~$\Real^m$ as
$\|\x\|_q \defin (\sum_{i=1}^m |\x[i]|^q)^{{1}/{q}}$, where~$\x[i]$ denotes the
$i$-th entry of~$\x$, and $\|\x\|_\infty \defin \max_{i=1,\ldots,m} |\x[i]|
= \lim_{q \to \infty} \|\x\|_q$.  We also define the $\ell_0$-pseudo-norm as
the number of nonzero elements in a vector.
We consider the
Frobenius norm of a matrix~$\X$ in~$\Real^{m \times n}$: $\|\X\|_\FFF \defin
(\sum_{i=1}^m \sum_{j=1}^n \X[i,j]^2)^{{1}/{2}}$.  We also write for a sequence
of vectors $\x_t$ and scalars~$u_t$, $\x_t = \O(u_t)$, when there exists a
constant $K > 0$ independent of $t$ so that for all $t$, $\|\x_t\|_2 \leq K
u_t$, and use a similar notation for matrices (note that for
finite-dimensional vector spaces, the choice of norm is irrelevant).
When $\Lambda \subseteq \{1,\ldots,m\}$ is a finite set of indices,
$\x_\Lambda$ denotes the vector in $\Real^{|\Lambda|}$ that carries the entries of
$\x$ indexed by $\Lambda$. Similarly, when $\X$ is a
matrix in $\Real^{m \times n}$ and $\Lambda \subseteq \{1,\ldots,n\}$,
$\X_\Lambda$ is the matrix in $\Real^{m \times |\Lambda|}$ whose columns 
are those of $\X$ indexed by $\Lambda$.

The rest of this paper is organized as follows: Section \ref{sec:classical}
presents the data-driven dictionary learning framework. Section
\ref{sec:taskdict} is devoted to our new task-driven framework, and Section~\ref{sec:optim} to efficient algorithms to addressing the corresponding
optimization problems. Section \ref{sec:exp} presents several dictionary learning experiments for
signal classification, signal regression, and compressed sensing.

\section{Data-Driven Dictionary Learning} \label{sec:classical}
Classical dictionary learning techniques~\cite{field,engan,aharon} consider a finite training set of
signals $\X=[\x_1,\ldots,\x_n]$ in~$\Real^{m \times n}$ and minimize the
empirical cost function
\begin{displaymath}
   g_n(\D) \defin \frac{1}{n} \sum_{i=1}^n \l_u(\x_i,\D),
\end{displaymath}
with respect
to a dictionary~$\D$ in~$\Real^{m \times p}$, each column
representing a dictionary element.
$\l_u$ is
a loss function such that $\l_u(\x,\D)$ should be small if~$\D$ is ``good'' at
representing the signal~$\x$ in a sparse fashion. 
As emphasized by the index $u$ of $\l_u$, this optimization problem is \emph{unsupervised}.
As others~(see, e.g., \cite{aharon}), we define $\l_u(\x,\D)$ as
the optimal value of a sparse coding problem. We choose here the
elastic-net formulation of~\cite{zou}:
\begin{equation}
   \l_u(\x,\D) \defin \min_{\alphab \in \Real^p}
   \frac{1}{2}\|\x-\D\alphab\|_2^2 + \lambda_1 \|\alphab\|_1 +
   \frac{\lambda_2}{2}\|\alphab\|_2^2,\label{eq:ell1penalty}
\end{equation}
where~$\lambda_1$ and~$\lambda_2$ are regularization parameters.
When~$\lambda_2=0$, this leads to the $\ell_1$ sparse decomposition problem, also
known  as {\em basis pursuit}~\cite{chen}, or {\em Lasso}~\cite{tibshirani}.
Here, our choice of the elastic-net formulation over the Lasso is mainly for stability reasons.
Using a parameter $\lambda_2 > 0$ makes the problem of
Eq.~(\ref{eq:ell1penalty}) strongly convex and, as shown later in this paper,
ensures its unique solution to be Lipschitz with respect to~$\x$ and~$\D$
with a constant depending on~$\lambda_2$. 
Whereas the stability of this solution is not necessarily an issue when
learning a dictionary for a reconstruction task, it has turned out to be
empirically important in some of our experiments with other tasks.

To prevent the $\ell_2$-norm of~$\D$ from being arbitrarily large, which would lead to arbitrarily
small values of~$\alphab$, it is common to constrain its columns
$\d_1,\ldots,\d_p$ to have $\ell_2$-norms less than or equal to one.  We will
call $\DD$ the convex set of matrices satisfying this constraint:
\begin{equation}
   \DD \defin \{ \D \in \Real^{m \times p} \st \forall j \in \{1,\ldots,p\},~~
   \|\d_j\|_2 \leq 1 \}. \label{eq:C}
\end{equation}
As pointed out by Bottou and Bousquet~\cite{bottou}, one is usually not interested in a perfect
minimization of the \emph{empirical} cost~$g_n(\D)$, but instead in the
minimization with respect to~$\D$ of the {\em expected} cost
\begin{equation}
   g(\D) \defin \E_{\x}[\l_u(\x,\D)] \stackrel{\text{a.s.}}{=} \lim_{n \to \infty} g_n(\D), \label{eq:datadict}
\end{equation}
where the expectation is taken relative to the (unknown) probability
distribution~$p(\x)$ of the data, and is supposed to be finite.\footnote{We use
``a.s.'' (almost surely) to denote convergence with probability one.} In
practice, dictionary learning problems often involve a large amount of data.
For instance when the vectors $\x$ represent image patches, $n$ can be up to
several millions in a single image. In this context, online learning techniques
have shown to be very efficient for obtaining a stationary point of this
optimization problem~\cite{mairal7}. In this paper, we propose to minimize an expected cost corresponding to a \emph{supervised} dictionary
learning formulation, which we now present.
\section{Proposed Formulation} \label{sec:taskdict}
We introduce in this section a general framework for learning dictionaries
adapted to specific supervised tasks, e.g., classification, as opposed to the
unsupervised formulation of the previous section, and present different
extensions along with possible applications. 
\subsection{Basic Formulation}
Obtaining a good performance in classification tasks is often related to the
problem of finding a good data representation. Sparse decompositions obtained
with data-driven learned dictionaries have been used for that purpose in
\cite{grosse} and \cite{raina}, showing promising results for audio data and
natural images. We present in this section a formulation for learning a
dictionary in a \emph{supervised} way for regression or classification tasks.

Given a dictionary~$\D$ obtained using the approach presented in the previous
section, a vector~$\x$ in~$\XX \subseteq \Real^m$ can be
represented as a sparse vector~$\alphab^\star(\x,\D)$, defined as the solution
of an elastic-net problem~\cite{zou}:
\begin{equation}
   \alphab^\star(\x,\D) \defin \argmin_{\alphab \in \Real^p}
   \frac{1}{2}\|\x-\D\alphab\|_2^2 + \lambda_1 \|\alphab\|_1 +
   \frac{\lambda_2}{2} \|\alphab\|_2^2. \label{eq:elas}
\end{equation}

We now assume that each signal~$\x$ in~$\XX$ is associated to 
a variable~$\y$ in~$\YY$, which we want to predict from~$\x$.
Concretely, the set~$\YY$ can either be \emph{a finite set of labels} in
classification tasks, or a \emph{subset of $\Real^q$ for some integer~$q$} in
regression tasks. 
We can now use the sparse vector~$\alphab^\star(\x,\D)$ as a feature representation of a
signal~$\x$ in a classical expected risk minimization formulation:
\begin{equation}
   \min_{\W \in \WW} f(\W) + \frac{\nu}{2}\|\W\|_\FFF^2,\label{eq:unsup}
\end{equation}
where $\W$ are model parameters which we want to learn, $\WW$ is a convex set,
$\nu$ is a regularization parameter, and~$f$ is a convex function defined as
\begin{equation}
 f(\W) \defin \E_{\y,\x} [\l_s\big(\y,\W,\alphab^\star(\x,\D)\big)].
\end{equation}
In this equation, $\l_s$ is a convex loss function that measures how well
one can predict~$\y$ by observing~$\alphab^\star(\x,\D)$ given the
model parameters~$\W$. For instance, it can be the square, logistic, or hinge loss from support vector
machines~(see~\cite{shawe}).
The index~$s$ of~$\l_s$ indicates here that the loss is
adapted to a \emph{supervised} learning problem.
The expectation is taken with respect to the unknown probability
distribution~$p(\y,\x)$ of the data. 
So far, the dictionary~$\D$ is be obtained in 
an unsupervised way. However, Mairal et al.~\cite{mairal6}, and Bradley and Bagnell~\cite{bradley} have
shown that better results can be achieved when the dictionary is obtained in a
fully supervised setting, tuned for the prediction task. We now introduce the
task-driven dictionary learning formulation, that consists of \emph{jointly}
learning~$\W$ and~$\D$ by solving 
\begin{equation}
   \min_{\D \in \DD, \W \in \WW} f(\D,\W) + \frac{\nu}{2}\|\W\|_\FFF^2,
   \label{eq:formulation} 
\end{equation}
where $\DD$ is a set of constraints defined in Eq.~(\ref{eq:C}), and $f$ has
the form
\begin{equation}
   f(\D,\W) \defin \E_{\y,\x} [\l_s\big(\y,\W,\alphab^\star(\x,\D)\big)]. \label{eq:deff}
\end{equation}
The main difficulty of this optimization problem comes from the
non-differentiability of $\alphab^\star$,
which is the solution of a nonsmooth optimization problem (\ref{eq:elas}).
Bradley and Bagnell~\citet{bradley} have tackled this difficulty by introducing
a smooth approximation of the sparse regularization which leads to smooth solutions, allowing the use
of implicit differentiation to compute the gradient of the cost function they have introduced.  This
approximation encourages some coefficients in $\alphab^\star$ to be small, and does
not produce true zeros. It can be used when ``true'' sparsity is not
required.  In a different formulation, Mairal et al.~\citet{mairal6} have used
nonsmooth sparse regularization, but used heuristics to tackle the optimization
problem.  We show in Section \ref{sec:optim} that better optimization tools
than these heuristics can be used, while keeping a nonsmooth regularization for
computing $\alphab^\star$.

A difference between supervised and unsupervised dictionary learning
is that overcompleteness---that is, the dictionaries have
more elements than the signal dimension, has not empirically
proven to be necessary. It is indeed often advocated for image processing
applications that having~$p > m$ provides better reconstruction
results~\cite{elad,mairal8}, but for discriminative tasks, perfect reconstruction
is not always required as long as discriminative features are captured
by the sparse coding procedure.

Minor variants of the formulation~(\ref{eq:formulation}) can also be considered: Non-negativity
constraints may be added on~$\alphab^\star$ and~$\D$, leading to a supervised
version of nonnegative matrix factorization~\citep{lee2},
regularized with a sparsity-inducing penalty. The function~$\l_s$ could also
take extra arguments such as~$\D$ and~$\x$ instead of just $\y,\W,\alphab^\star$. For
simplicity, we have omitted these possibilities, but the formulations and
algorithms we present in this paper can easily be extended to these cases.

Before presenting extensions and applications of the formulation we have introduced, let us first discuss 
the assumptions under which our analysis holds.
\subsubsection{Assumptions} 
From now on, we assume that:
\begin{enumerate}
   \item[\bf (A)] The data $(\y,\x)$ admits a probability density~$p$ with
      a compact support $K_{\YY} \times K_{\XX} \subseteq \YY \times \XX$. This
      is a reasonable assumption in audio, image, and video processing
      applications, where it is imposed by the data acquisition process,
      where values returned by sensors are bounded. To simplify the notation
      we assume from now on that $\XX$ and $\YY$ are
      compact.\footnote{Even though images are acquired in practice after a quantization process, it is a common assumption in image processing to consider pixel values in a continuous space.}
   \item[\bf (B)] When $\YY$ is a subset of a finite-dimensional real vector space, 
      $p$ is continuous and~$\l_s$ is twice continuously differentiable.
   \item[\bf (C)] When $\YY$ is a finite set of labels, for all~$\y$ in~$\YY$,
      $p(\y,.)$ is continuous and $\l_s(\y,.)$ is twice continuously
      differentiable.\footnote{For a given value of~$\y$ and function $g$, $g(\y,.)$ denotes
      the function which associates to a vector~$\x$ the value~$g(\y,\x)$.
      }
\end{enumerate}
Assumptions {\bf (B)} and {\bf (C)} allow us to use several loss functions such
as the square, logistic, or softmax losses. 

\subsection{Extensions} \label{subsec:extensions}
We now present two extensions of the previous formulations. The first one
includes a linear transform of the input data, and the second one exploits
unlabeled data in a semi-supervised setting.
\subsubsection{Learning a Linear Transform of the Input Data}
\label{subsec:linear}
In this section, we add to our basic formulation a linear transform of the
input features, represented by a matrix~$\Z$. Our motivation for this 
is twofold: It can be appealing to reduce the dimension of the feature space
via such a linear transform, and/or it can make the model richer by
increasing the numbers of free parameters. The resulting formulation
is the following:
\begin{equation}
   \min_{\D \in \DD, \W \in \WW, \Z \in \ZZ} f(\D,\W,\Z) +\frac{\nu_1}{2}\|\W\|_\FFF^2 + \frac{\nu_2}{2}\|\Z\|_\FFF^2,\label{eq:extended}
\end{equation}
where $\nu_1$ and $\nu_2$ are two regularization parameters, $\ZZ$ is a convex set and
\begin{equation}
   f(\D,\W,\Z) \defin \E_{\y,\x} [\l_s\big(\y,\W,\alphab^\star(\Z\x,\D)\big)]. \label{eq:generalf}
\end{equation}
It is worth noticing that the formulations of
Eq.~(\ref{eq:formulation}) and Eq.~(\ref{eq:extended}) can also be extended to the
case of a cost function depending on several dictionaries involving several
sparse coding problems, such as the one used in~\citet{mairal3} for signal
classification.  Such a formulation is not developed here for simplicity
reasons, but algorithms to address it can easily be derived from this paper.

\subsubsection{Semi-supervised Learning} \label{subsec:semi}
As shown in~\citet{raina}, sparse coding techniques can be effective for
learning good features from unlabeled data. The extension of our task-driven
formulation to the semi-supervised learning setting is natural and takes
the form 
\ifthenelse{\isundefined{\IEEEcompilation}}{
\begin{equation}
    \min_{\D \in \DD, \W \in \WW} (1-\mu)\E_{\y,\x} [\l_s\big(\y,\W,\alphab^\star(\x,\D)\big)] + 
     \mu \E_{\x} [\l_u(\x,\D)]  + \frac{\nu}{2}\|\W\|_\FFF^2, \label{eq:semi}
 \end{equation}
}{
\begin{multline}
    \min_{\D \in \DD, \W \in \WW} (1-\mu)\E_{\y,\x} [\l_s\big(\y,\W,\alphab^\star(\x,\D)\big)] + \\
    \mu \E_{\x} [\l_u(\x,\D)]  + \frac{\nu}{2}\|\W\|_\FFF^2, \label{eq:semi}
\end{multline}
}
where the second expectation is taken with respect to the marginal distribution of $\x$.
The function~$\l_u$ is the loss function defined in Eq.~(\ref{eq:ell1penalty}), and $\mu$ in
$[0,1]$ is a new parameter controlling the trade-off between the
unsupervised and supervised learning cost functions.

\subsection{Applications} \label{subsec:applications}
For illustration purposes, we present a few applications of our
task-driven dictionary learning formulations. 
Our approach
is of course not limited to these examples.
\subsubsection{Regression} \label{subsec:regression}
In this setting, $\YY$ is a subset of a $q$-dimensional real vector space, and
the task is to predict variables~$\y$ in~$\YY$ from the observation of vectors~$\x$
in~$\XX$.  A typical application is for instance the restoration of clean signals~$\y$ from
observed corrupted signals~$\x$. Classical signal restoration techniques often focus on
removing additive noise or solving inverse linear problems~\citep{daubechies}.
When the corruption results from an unknown nonlinear transformation, we
formulate the restoration task as a general regression problem.
This is the case for example in the experiment presented in
Section~\ref{subsec:expartefacts}.

We define the task-driven dictionary learning formulation for regression as
follows:
\begin{equation}
   \min_{\W \in \WW, \D \in \DD} \E_{\y,\x}
   \Big[\frac{1}{2}\|\y-\W\alphab^\star(\x,\D)\|_2^2\Big]  +
   \frac{\nu}{2}\|\W\|_\FFF^2. \label{eq:regression}
\end{equation}
At test time, when a new signal~$\x$ is observed, the estimate of the corresponding
variable~$\y$ provided by this model is $\W\alphab^\star(\x,\D)$ (plus possibly
an intercept which we have omitted here for simplicity reasons). Note that we
here propose to use the square loss for estimating the difference between $\y$
and its estimate $\W\alphab^\star(\x,\D)$, but any other twice differentiable
loss can be used.

\subsubsection{Binary Classification} \label{subsec:binary}
In this section and in the next one, we propose to learn dictionaries adapted
to classification tasks. Our approach follows the formulation presented
in~\citep{mairal6}, but is slightly different and falls into our task-driven 
dictionary learning framework.
In this setting, the set~$\YY$ is equal to $\{-1; +1\}$.
Given a vector~$\x$, we want to learn the parameters~$\w$
in~$\Real^p$ of a linear model to predict~$y$ in~$\YY$, using the sparse
representation $\alphab^\star(\x,\D)$ as features, and jointly optimize~$\D$ and
$\w$. For instance, using the logistic regression loss, our formulation becomes
\begin{equation}
   \min_{\w \in \Real^p,\D \in \DD} \E_{y,\x}
   \Big[\log\big(1+e^{-y \w^\top \alphab^\star(\x,\D)}\big)\Big]  +
   \frac{\nu}{2}\|\w\|_2^2,  \label{eq:linear}
\end{equation}
Once $\D$ and $\w$ have been learned, a new signal~$\x$ is classified
according to the sign of $\w^\top \alphab^\star(\x,\D)$.
For simplicity reasons, we have omitted the intercept in the linear model,
but it can easily be included in the formulation. Note that 
instead of the logistic regression loss, any other twice
differentiable loss can be used.

As suggested in \citet{mairal6}, it is possible to extend this approach
with a bilinear model by learning a matrix~$\W$ so that a new vector~$\x$ is classified
according to the sign of $\x^\top \W \alphab^\star(\x,\D)$. In 
this setting, our formulation becomes
\begin{equation}
   \min_{\W \in \Real^{m \times p},\D \in \DD} \E_{y,\x}
   \Big[\log\big(1+e^{-y \x^\top \W \alphab^\star(\x,\D)}\big)\Big]  +
   \frac{\nu}{2}\|\W\|_\FFF^2.  \label{eq:bilinear}
\end{equation}
This bilinear model requires learning~$pm$ parameters as opposed to the~$p$
parameters of the linear one.  It is therefore richer and can sometimes offer a better
classification performance when the linear model is not rich enough to explain
the data, but it might be more subject to overfitting.

Note that we have naturally presented the binary classification task using the
logistic regression loss, but as we have experimentally observed, the square loss is also
an appropriate choice in many situations.

\subsubsection{Multi-class Classification}
When~$\YY$ is a finite set of labels in $\{1,\ldots,q\}$ with \mbox{$q > 2$},
extending the previous formulation to the multi-class setting can be done in
several ways, which we briefly describe here. 
The simplest possibility is to use a set of binary classifiers presented in
Section~\ref{subsec:binary} in a ``one-vs-all'' or ``one-vs-one'' scheme.  Another
possibility is to use a multi-class cost function such as the soft-max
function, to find linear predictors $\w_k$, $k$ in $\{1,\ldots,q\}$ 
such that for a vector $\x$ in $\XX$, the quantities $\w_y^\top
\alphab^\star(\x,\D)$ are encouraged to be greater than $\w_k^\top
\alphab^\star(\x,\D)$ for all $k \neq y$.
Another possibility is to turn the multi-class classification problem into a regression one
and consider that $\YY$ is a set of $q$ binary vectors of
dimension $q$ such that the $k-$th vector has $1$ on its
$k$-th coordinate, and $0$ elsewhere. This allows using the 
regression formulation of Section~\ref{subsec:regression} to solve the classification problem.

We remark that for classification tasks, scalability issues should be
considered when choosing between a one-vs-all scheme (learning independent
dictionaries for every class) and using a multi-class loss function (learning a
single dictionary shared between all classes).  The one-vs-all scheme requires
keeping into memory $qpm$ parameters, where~$q$ is the number of classes, which
is feasible when $q$ is reasonably small. For classifications problems with
many classes (for instance $q \geq 1\,000$), using a single (larger) dictionary
and a multi-class loss function is more appropriate, and would in addition
allow feature sharing between the classes.

\subsubsection{Compressed sensing} \label{subsec:cs}
Let us consider a signal~$\x$ in~$\Real^{m}$, the theory of compressed
sensing~\cite{candes3,donoho3} tells us that under certain 
assumptions, the vector~$\x$ can be recovered exactly from a few
measurements~$\Z\x$, where~$\Z$ in~$\Real^{r \times m}$ is called a ``sensing''
matrix with~$r \ll m$. Unlike classical signal processing methods, such a
linear transformation is sometimes included physically in the data acquisition
process itself~\citep{duarte2}, meaning that a sensor can provide measurements~$\Z\x$ 
without directly measuring~$\x$.

In a nutshell, the recovery of~$\x$ has been proven to be possible 
when~$\x$ admits a sparse representation on a
dictionary~$\D$, and the sensing matrix~$\Z$ is incoherent with~$\D$, meaning
that the rows of~$\Z$ are sufficiently uncorrelated with the columns of~$\D$
(see~\citealp{candes3,donoho3} for more details).\footnote{The
assumption of ``incoherence'' between~$\D$ and~$\Z$ can be replaced with a
different but related hypothesis called \emph{restricted isometry property}.
Again the reader should refer to \citep{candes3,donoho3} for more details.} To
ensure that this condition is satisfied, $\Z$ is often chosen as a random matrix,
which is incoherent with any dictionary with high probability.

The choice of a random matrix is appealing for many reasons. In addition
to the fact that it provides theoretical guarantees of incoherence, it is well
suited to the case where~$m$ is large, making it impossible to store a
deterministic matrix~$\Z$ into memory, whereas it is sufficient to store
the seed of a random process to generate a random matrix.  On the other hand, large signals can
often be cut into smaller parts that still admit sparse decompositions, e.g.,
image patches, which can be treated independently with a deterministic smaller
matrix~$\Z$. When this is the case or when~$m$ has a reasonable size,
the question of whether to use a deterministic matrix~$\Z$ or
a random one arises, and it has been empirically observed that learned
matrices~$\Z$ can outperform random projections: For example, it is shown in~\citet{weiss}
that classical dimensionality reduction techniques such as
principal component analysis (PCA) or independent component analysis (ICA)
could do better than random projections in noisy settings, and in~\citet{duarte}
that jointly learning sensing matrices and dictionaries can
do even better in certain cases. A Bayesian framework for
learning sensing matrices in compressed sensing applications is also
proposed in \citet{seeger2}.

Following the latter authors, we study the case where~$\Z$ is not random
but learned at the same time as the dictionary, and introduce a 
formulation which falls into out task-driven dictionary learning framework:
\begin{equation}
   \min_{\substack{\D \in \DD \\ \W \in \Real^{m \times p} \\ \Z \in \Real^{r \times m}}} \E_{\y,\x}
   \Big[\frac{1}{2}\|\y-\W\alphab^\star(\Z\x,\D)\|_2^2\Big] + 
   \frac{\nu_1}{2}\|\W\|_\FFF^2+\frac{\nu_2}{2}\|\Z\|_\FFF^2,\label{eq:cs}
\end{equation}
where we learn $\D$, $\W$ and $\Z$ so that the variable~$\y$ should be well
reconstructed when encoding the ``sensed'' signal~$\Z\x$ with a dictionary
$\D$. In a noiseless setting,~$\y$ is naturally set to the same value as~$\x$.
In a noisy setting, it can be a corrupted version of~$\x$.

After having presented our general task-driven dictionary learning formulation, we present
next a strategy to address the corresponding nonconvex optimization problem.
\section{Optimization} \label{sec:optim}
We first show that the cost function $f$ of our basic
formulation~(\ref{eq:formulation}) is differentiable and compute its gradient.
Then, we refine the analysis for the different variations presented in the
previous section, and describe an efficient online learning algorithm to
address them.

\subsection{Differentiability of $f$}
We analyze the differentiability of~$f$ as defined in Eq.~(\ref{eq:formulation})
with respect to its two arguments $\D$ and~$\W$. We consider here the case
where $\YY$ is a compact subset of a finite dimensional real vector space, but all proofs and
formulas are similar when $\YY$ is a finite set of labels.  The purpose of this
section is to show that even though the sparse coefficients $\alphab^\star$ are
obtained by solving a non-differentiable optimization problem, $f$ is
differentiable on $\WW \times \DD$, and one can compute its gradient.

The main argument in the proof of Propositions~\ref{prop:regf}
and~\ref{prop:regfext} below is that, although the
function~$\alphab^\star(\x,\D)$ is not differentiable, it is uniformly
Lipschitz continuous, and differentiable almost everywhere.  The only points
where $\alphab^\star$ is not differentiable are points where the set
of nonzero coefficients of $\alphab^\star$ change (we always denote this set by $\Lambda$
in this paper).  Considering optimality conditions of the elastic-net
formulation of Eq.~(\ref{eq:ell1penalty}), these points are easy to characterize.
The details of the proof have been relegated to the Appendix (Lemma~\ref{lemma:elas} and Proposition~\ref{prop:elasreg}) for
readability purposes. With these results in hand, we then show that $f$ admits a first-order
Taylor expansion meaning that it is differentiable, the sets where $\alphab^\star$
is not differentiable being negligible in the expectation from 
the definition of $f$ in Eq.~(\ref{eq:deff}).
We can now state our main result:
\UpperSpace
\begin{proposition}[Differentiability and gradients of $f$] \label{prop:regf} ~\newline
   Assume $\lambda_2 >0$, ({\bf A}), ({\bf B}) and ({\bf C}). Then, the
   function $f$ defined in Eq.~(\ref{eq:formulation}) is differentiable, and
   \begin{equation}
      \left\{
      \begin{aligned}
         \nabla_\W f(\D,\W) & = \E_{\y,\x}[\nabla_\W \l_s(\y,\W,\alphab^\star)], \\
         \nabla_\D f(\D,\W) & = \E_{\y,\x}[-\D\betab^\star\alphab^{\star\top} + (\x-\D\alphab^\star)\betab^{\star\top}], \\
      \end{aligned}  
      \right.
      \label{eq:gradient}
   \end{equation}
   where $\alphab^\star$ is short for $\alphab^\star(\x,\D)$, 
   and $\betab^\star$ is a vector in $\Real^p$ that depends on $\y,\x,\W,\D$ with
   \begin{equation}
      \betab^\star_{\Lambda^C}  = 0 ~~\text{and}~~ \betab^\star_{\Lambda}  = (\D_\Lambda^\top \D_\Lambda+\lambda_2 \I)^{-1} \nabla_{\alphab_\Lambda} \l_s(\y,\W,\alphab^\star), 
      \label{eq:beta}
   \end{equation}
   where $\Lambda$ denotes the indices of the nonzero coefficients of $\alphab^\star(\x,\D)$.
\end{proposition}
\LowerSpace
The proof of this proposition is given in Appendix. 
We have shown that the function defined in Eq.~(\ref{eq:formulation}) is smooth, and computed its gradients.
The same can be done for the more general formulation of Eq.~(\ref{eq:generalf}):
\UpperSpace
\begin{proposition}[Differentiability, extended formulation] \label{prop:regfext} ~\newline
   Assume $\lambda_2 >0$, ({\bf A}), ({\bf B}) and ({\bf C}). Then, the function $f$ defined in Eq.~(\ref{eq:generalf}) is
   differentiable. The gradients of $f$ are 
   \begin{equation}
      \left\{
      \begin{aligned}
         \nabla_\W f(\D,\W,\Z) & = \E_{\y,\x}[\nabla_\W \l_s(\y,\W,\alphab^\star)], \\
         \nabla_\D f(\D,\W,\Z) & = \E_{\y,\x}[-\D\betab^\star\alphab^{\star\top} + (\Z\x-\D\alphab^\star)\betab^{\star\top}], \\
         \nabla_\Z f(\D,\W,\Z) & = \E_{\y,\x}[\D\betab^\star\x^\top], \\
      \end{aligned}  
      \right.
      \label{eq:gradient2}
   \end{equation}
   where $\alphab^\star$ is short for $\alphab^\star(\Z\x,\D)$, 
   and $\betab^\star$ is defined in Eq.~(\ref{eq:beta}).
\end{proposition}
\LowerSpace
The proof is similar to the one of Proposition~\ref{prop:regf} in Appendix, and uses similar arguments.

\subsection{Algorithm} \label{subsec:algo}
Stochastic gradient descent algorithms are typically designed to
minimize functions whose gradients have the form of an expectation as in Eq.~(\ref{eq:gradient}).
They have been shown to converge to stationary points of (possibly nonconvex)
optimization problems under a few assumptions that are a bit stricter than the
ones satisfied in this paper~(see~\citep{bottou} and references
therein).\footnote{As often done in machine learning, we use stochastic gradient descent in a setting
where it is not guaranteed to converge in theory, but is has proven to behave
well in practice, as shown in our experiments. The convergence proof of Bottou~\cite{bottou} for non-convex problems indeed assumes three times differentiable cost functions.}
As noted in~\citet{mairal7}, these
algorithms are generally well suited
to unsupervised dictionary learning when their learning rate is well tuned.

The method we propose here is a projected first-order stochastic gradient
algorithm~(see \citep{kushner}), and it is given in Algorithm~\ref{algo:sgd}.
It sequentially draws i.i.d samples $(\y_t,\x_t)$ from the probability
distribution $p(\y,\x)$. Obtaining such i.i.d. samples may be difficult since
the density $p(\y,\x)$ is unknown. At first approximation, the vectors $(\y_t,\x_t)$ are obtained
in practice by cycling over a randomly permuted training set, which is often
done in similar machine learning settings~\citep{bottou2}.

\begin{algorithm}[hbtp]
   \caption{Stochastic gradient descent algorithm for task-driven dictionary learning.}
   \label{algo:sgd}
   \begin{algorithmic}[1]
      \REQUIRE $p(\y,\x)$ (a way to draw i.i.d samples of~$p$), $\lambda_1,\lambda_2,\nu \in \Real$ (regularization
      parameters), $\D \in \DD$ (initial dictionary), $\W \in \WW$ (initial parameters), $T$
      (number of iterations), $t_0,\rho$ (learning rate parameters).
      \FOR {$t= 1$ to $T$}
      \STATE Draw $(\y_t,\x_t$) from $p(\y,\x)$. 
      \STATE Sparse coding: compute $\alphab^\star$ using a modified LARS~\cite{efron}. 
      \begin{displaymath} 
         \alphab^\star  \leftarrow  \argmin_{\alphab \in
         \Real^p} \frac{1}{2}||\x_t-\D\alphab||_2^2 +
         \lambda_1||\alphab||_1 + \frac{\lambda_2}{2}||\alphab||_2^2.
      \end{displaymath}
      \STATE Compute the active set: $$\Lambda \leftarrow \{ j \in \{1,\ldots,p\} : \alphab^\star[j] \neq 0 \}.$$
      \STATE Compute $\betab^\star$: Set $ \betab^\star_{\Lambda^C}  = 0$ and 
      \begin{displaymath}
         \betab^\star_{\Lambda}  = (\D_\Lambda^\top \D_\Lambda+\lambda_2 \I)^{-1} \nabla_{\alphab_\Lambda} \l_s(\y_t,\W,\alphab^\star).
      \end{displaymath}
      \STATE Choose the learning rate $\rho_t \leftarrow \min\big(\rho,\rho\frac{t_0}{t}\big)$.
      \STATE Update the parameters by a projected gradient step
      \begin{displaymath}
         \begin{split}
            \W &\leftarrow \Pi_{\WW}\Big[\W - \rho_t \big(\nabla_\W \l_s(\y_t,\W,\alphab^\star) + \nu \W \big) \Big], \\
            \D &\leftarrow \Pi_{\DD}\Big[\D - \rho_t \big( -\D\betab^\star\alphab^{\star\top} + (\x_t-\D\alphab^\star)\betab^{\star\top}\big)\Big], \\
         \end{split}
      \end{displaymath}
      where $\Pi_{\WW}$ and $\Pi_{\DD}$ are respectively orthogonal projections on the sets~$\WW$ and~$\DD$.
   \ENDFOR
   \RETURN $\D$ (learned dictionary).
\end{algorithmic}
\end{algorithm}

At each iteration, the sparse code $\alphab^\star(\x_t,\D)$ is computed by
solving the elastic-net formulation of~\citet{zou}.  We have chosen to use the
LARS algorithm, a homotopy method~\cite{efron}, which was originally
developed to solve the Lasso formulation---that is, $\lambda_2=0$, but which
can be modified to solve the elastic-net problem.  Interestingly, it admits an
efficient implementation that provides a Cholesky decomposition of the matrix
$(\D_\Lambda^\top \D_\Lambda+\lambda_2\I)^{-1}$~(see \cite{zou,efron}) as
well as the solution~$\alphab^\star$.  In this setting, $\betab^\star$ can be
obtained without having to solve from scratch a new linear system.

The learning rate $\rho_t$ is chosen according to a heuristic rule. Several
strategies have been presented in the literature~(see \cite{lecun2,murata} and
references therein). A classical setting uses a
learning rate of the form $\rho / t$, where~$\rho$ is a constant.\footnote{A $1/t$-asymptotic learning rate is usually used for proving the convergence of stochastic gradient descent algorithms~\cite{bottou}.}
However, such a learning rate is known to decrease too quickly in many practical
cases, and one sometimes prefers a learning rate of the form $\rho / (t+t_0)$,
which requires tuning two parameters. In this paper, we have chosen a learning
rate of the form $\min(\rho,\rho t_0/t)$---that is, a constant learning rate
$\rho$ during $t_0$ iterations, and a $1/t$ annealing strategy when $t > t_0$,
a strategy used by~\citet{murata} for instance. Finding good parameters~$\rho$ 
and~$t_0$ also requires in practice a good heuristic. The one we have used
successfully in all our experiments is $t_0=T/10$, where $T$ is the total
number of iterations. Then, we try several values of $\rho$ during a few
hundreds of iterations and keep the one that gives the lowest error on a small
validation set.

In practice, one can also improve the convergence speed of our algorithm with a
mini-batch strategy---that is, by drawing $\eta > 1$ samples at each iteration instead of
a single one. This is a classical heuristic in stochastic gradient descent
algorithms and, in our case, this is further motivated by the fact that solving
$\eta$ elastic-net problems with the same dictionary $\D$ can be accelerated by
the precomputation of the matrix $\D^\top\D$ when $\eta$ is large enough.  Such
a strategy is also used in~\citet{mairal7} for the classical data-driven
dictionary learning approach. In practice, the value $\eta=200$ has given good results in 
all our experiments (a value found to be good for the unsupervised setting as well).

As many algorithms tackling non-convex optimization problems, our method for
learning supervised dictionaries can lead to poor results if is not well
initialized. The classical unsupervised approach of dictionary
learning presented in Eq.~(\ref{eq:datadict}) has been found empirically to be
better behaved than the supervised one, and easy to
initialize~\citep{mairal7}. We therefore have chosen to initialize our
dictionary~$\D$ by addressing the unsupervised formulation of
Eq.~(\ref{eq:datadict}) using the SPAMS toolbox~\citep{mairal7}.\footnote{\url{http://www.di.ens.fr/willow/SPAMS/}} With this
initial dictionary~$\D$ in hand, we optimize with respect to $\W$ the cost
function of Eq~(\ref{eq:unsup}), which is convex. This procedure gives us
a pair $(\D,\W)$ of parameters which are used to initialize Algorithm~\ref{algo:sgd}.

\subsection{Extensions}
We here present the slight modifications to Algorithm~\ref{algo:sgd} necessary to address
the two extensions discussed in Section~\ref{subsec:extensions}.

The last step of Algorithm~\ref{algo:sgd} updates the parameters~$\D$ and~$\W$
according to the gradients presented in Eq.~(\ref{eq:gradient2}). Modifying the algorithm to address the formulation of
Section~\ref{subsec:linear} also requires updating the parameters $\Z$
according to the gradient from Proposition~\ref{prop:regfext}:
\begin{displaymath}
   \Z \leftarrow \Pi_{\ZZ}\Big[\Z - \rho_t (\D\betab^\star\x^{\top}+\nu_2\Z)\Big],
\end{displaymath}
where $\Pi_{\ZZ}$ denotes the orthogonal projection on the set~$\ZZ$.

The extension to the semi-supervised formulation of Section~\ref{subsec:semi}
assumes that one can draw samples from the marginal distribution $p(\x)$.  This
is done in practice by cycling over a randomly permuted set of unlabeled
vectors. Extending Algorithm~\ref{algo:sgd} to this setting requires the
following modifications: At every iteration, we draw one pair $(\y_t,\x_t)$
from $p(\y,\x)$ and one sample $\x'_{t}$ from~$p(\x)$.  We proceed exactly as
in Algorithm~\ref{algo:sgd}, except that we also compute
$\alphab^{\star\prime}\defin \alphab^\star(\x'_t,\D)$, and replace the update
of the dictionary~$\D$ by
\ifthenelse{\isundefined{\IEEEcompilation}}{
\begin{equation}
   \D \leftarrow \Pi_{\DD}\Big[\D - \rho_t \Big( (1-\mu) \big(
   -\D\betab^\star\alphab^{\star\top} +
   (\x_t-\D\alphab^\star)\betab^{\star\top}\big) + 
 \mu\big(
   -(\x'_t-\D\alphab^{\star\prime})\alphab^{\star\prime\top}\big) \Big) \Big],
\end{equation}
}{
\begin{multline}
   \D \leftarrow \Pi_{\DD}\Big[\D - \rho_t \Big( (1-\mu) \big(
   -\D\betab^\star\alphab^{\star\top} +
   (\x_t-\D\alphab^\star)\betab^{\star\top}\big) + \\
 \mu\big(
   -(\x'_t-\D\alphab^{\star\prime})\alphab^{\star\prime\top}\big) \Big) \Big],
\end{multline}
}
where the term $-(\x_t'-\D\alphab^{\star\prime})\alphab^{\star\prime\top}$ is
in fact the gradient $\nabla_\D \ell_u(\x_t,\D)$, as shown in \cite{mairal7}. 
\section{Experimental Validation} \label{sec:exp}
Before presenting our experiments, we briefly discuss the question of
choosing the parameters in our formulation. 
\subsection{Choosing the Parameters} \label{subsec:parameters}
Performing cross-validation on the parameters $\lambda_1$, $\lambda_2$ (elastic-net
parameters), $\nu$ (regularization parameter) and $p$ (size of the dictionary)
would of course be cumbersome, and we use a few simple heuristics to either
reduce the search space for these parameters or fix arbitrarily some of them.
We have proceeded in the following way:
\begin{itemize}
   \item Since we want to exploit sparsity, we often set~$\lambda_2$ to~$0$, even though $\lambda_2 > 0$ is necessary in our
      analysis for proving the differentiability of our cost function. This has
      proven to give satisfactory results in most experiments, except for the
      experiment of Section~\ref{sec:expcs}, where choosing a small positive value for
      $\lambda_2$ was necessary for our algorithm to converge.
   \item We have empirically observed that natural image patches (that are
      preprocessed to have zero-mean and unit $\ell_2$-norm) are usually well reconstructed
      with values of $\lambda_1$ around $0.15$ (a value used in \cite{mairal6}
      for instance), and that one only needs to test a few
      different values, for instance $\lambda_1 = 0.15 + 0.025k$, with $k \in \{-3,\ldots,3\}$.
   \item When there is a lot of training data, which is often the case for
      natural image patches, the regularization with $\nu$ becomes unnecessary
      and this parameter can arbitrarily set to a small value, e.g.,
      $\nu=10^{-9}$ for normalized input data.  When there are not many training
      points, this parameter is set up by cross-validation.
   \item We have also observed that a larger dictionary usually means a better performance,
      but a higher computational cost. Setting the size of the dictionary is therefore 
      often a trade-off between results quality and efficiency. In our
      experiments, we often try the values $p$ in $\{50,100,200,400\}$.
\end{itemize}
We show in this section several applications of our method to real problems,
starting with handwritten digits classification, then moving to the 
restoration of images damaged by an unknown
nonlinear transformation, digital art authentification, and compressed sensing. 

\subsection{Handwritten Digits Classification}
We consider here a classification task using the MNIST~\cite{lecun3} and USPS~\cite{lecun4}
handwritten datasets. MNIST contains $70\,000$ $28 \times 28$
images, $60\,000$ for training, $10\,000$ for testing,  whereas
USPS has $7\,291$ training
images and $2\,007$ test images of size $16\times16$.  

We address this multiclass classification problem with a one-vs-all strategy,
learning independently one dictionary and classifier per class, using the
formulation of Section~\ref{subsec:binary}. This approach has proven here to be
faster than learning a single large dictionary with a multi-class loss
function, while providing very good results.  In this experiment, the
Lasso~\citet{tibshirani} is preferred to the elastic-net
formulation~\citet{zou}, and $\lambda_2$ is thus set to~$0$.  All digits are
preprocessed to have zero-mean and are normalized to have unit $\ell_2$-norm.
For the reasons mentioned earlier, we try the parameters $\lambda_1 = 0.15 +
0.025k$, with $k \in \{-3,\ldots,3\}$, and and~$\nu$ is chosen in
$\{10^{-1},\ldots,10^{-6}\}$. We select the parameters on MNIST by keeping the
last $10\,000$ digits of the training set for validation, while training on the
first $50\,000$ ones.  For USPS, we similarly keep $10\%$ of
the training set for validation.  Note that a cross-validation scheme may
give better results, but would be computationally more expensive.

Most effective digit recognition techniques use features with
shift invariance properties~\cite{ranzato2,haasdonk}. Since our formulation is
less sophisticated than for instance the convolutional network architecture
of~\cite{ranzato2} and does not enjoy such properties, we have artificially
augmented the size of the training set by considering versions of the digits
that are shifted by one pixel in every direction.  This is of course not an
optimal way of introducing shift invariance in our framework, but it is fairly
simple.

After choosing the parameters using the validation set, we retrain
our model on the full training set.  Each experiment is performed with
$40\,000$ iterations of our algorithm with a mini-batch of size $200$. We report the performance on
the test set achieved for different dictionary sizes, with $p$ in
$\{50,100,200,300\}$ for the two datasets, and observe that learning $\D$ in a
supervised way significantly improves the performance of the classification.
Moreover our method achieves state-of-the-art results on MNIST with a $0.54\%$ error rate, which 
is similar to the $0.60\%$ error rate of~\citet{ranzato2}.\footnote{It is also
shown in \cite{ranzato2} that better results can be achieved by considering
deformations of the training set.} Our $2.84\%$ error rate on USPS is slightly
behind the $2.4\%$ error rate of~\citet{haasdonk}.

We remark that a digit recognition task was also carried out in~\cite{mairal6}, where a similar 
performance is reported.\footnote{The error rates in~\cite{mairal6} are slightly higher
but the dataset used in their paper is not augmented with shifted versions of the digits.}
Our conclusions about the advantages of supervised versus unsupervised
dictionary learning are consistent with~\cite{mairal6}, but our approach
has two main advantages. First it is much easier to use
since it does not requires complicated heuristic procedures to select the parameters, and second it applies to a wider spectrum of applications such as to regression tasks.
\begin{table}
   \renewcommand{\arraystretch}{1.3}
   \caption{Test error in percent of our method for the digit recognition task for different dictionary sizes $p$.}\label{table:digits}
   \centering
\ifthenelse{\isundefined{\IEEEcompilation}}{
\vspace*{0.25cm}
}{ }
   \begin{tabular}{|c||c|c|c|c||c|c|c|c|}
      \hline
      $\D$ & \multicolumn{4}{c||}{unsupervised} & \multicolumn{4}{c|}{supervised} \\
      \hline
      $p$ & $50$ & $100$ & $200$ & $300$  & $50$ & $100$ & $200$ & $300$   \\
      \hline
      MNIST & 5.27 & 3.92 & 2.95 & 2.36 & .96 &  .73 &  .57 &  .54 \\
      \hline
      USPS & 8.02 &  6.03 & 5.13 & 4.58 & 3.64 & 3.09 & 2.88 & 2.84 \\
      \hline
   \end{tabular}
\end{table}

\begin{figure}
   \centering
   \includegraphics[width=0.9\linewidth]{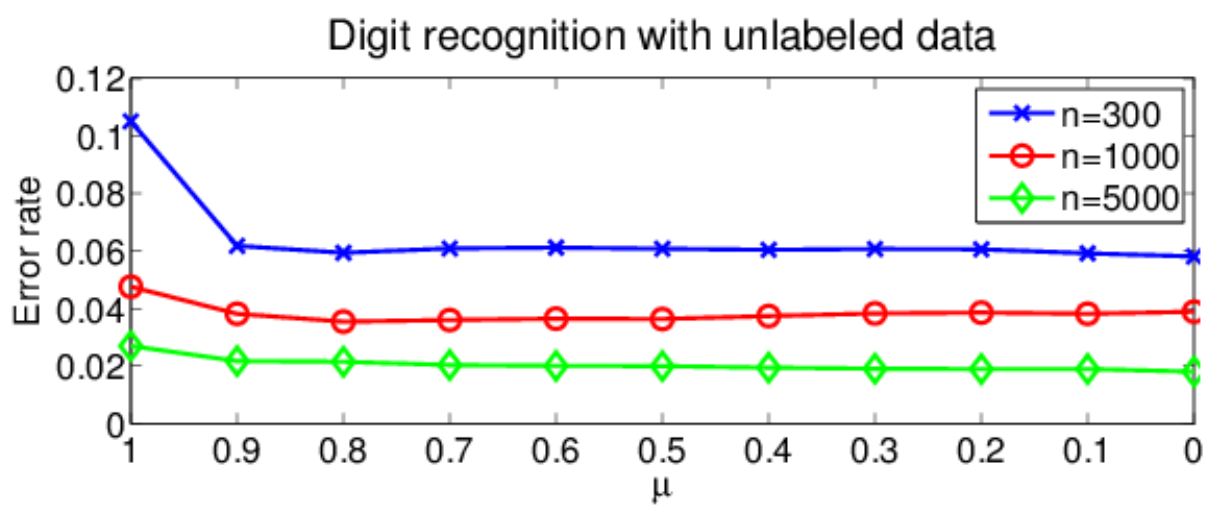} \hfill
   \caption{Error rates on MNIST when using $n$ labeled data, for various values of $\mu$.}
\label{fig:unlabeled}
\end{figure}

Our second experiment follows~\citet{ranzato2},
where only a few samples are labelled.
We use the semi-supervised formulation of
Section~\ref{subsec:semi} which exploits unlabeled data.
Unlike the first experiment where the parameters
are chosen using a validation set, and following~\citet{ranzato2}, we make a
few arbitrary choices. Indeed, we use $p=300$,
$\lambda_1=0.075$, and $\nu=10^{-5}$, which were the parameters
chosen in the previous experiment.
As in the previous experiment, we have observed that these parameters lead to sparse
vectors~$\alphab^\star$ with about~$15$ non-zero coefficients.
The dictionaries associated with each digit class are
initialized using the unsupervised formulation of Section~\ref{sec:classical}.
To test our algorithm with different values of $\mu$, we
use a continuation strategy: Starting with $\mu=1.0$, we sequentially decrease
its value by $0.1$ until we have $\mu=0$, learning with $10\,000$ iterations 
for each new value of $\mu$.
We report the error rates in Figure~\ref{fig:unlabeled},
showing that our approach offers a competitive performance similar to~\citet{ranzato2}. The best error rates of our method for
$n=300,1000,5000$ labeled data are respectively $5.81, 3.55$ and $1.81 \%$, which is similar to
\citet{ranzato2} who has reported $7.18, 3.21$ and $1.52\%$ with the same sets of labeled data.

\subsection{Learning a Nonlinear Image Mapping}\label{subsec:expartefacts}
\begin{figure}
   \centering
    \includegraphics[width=\linewidth]{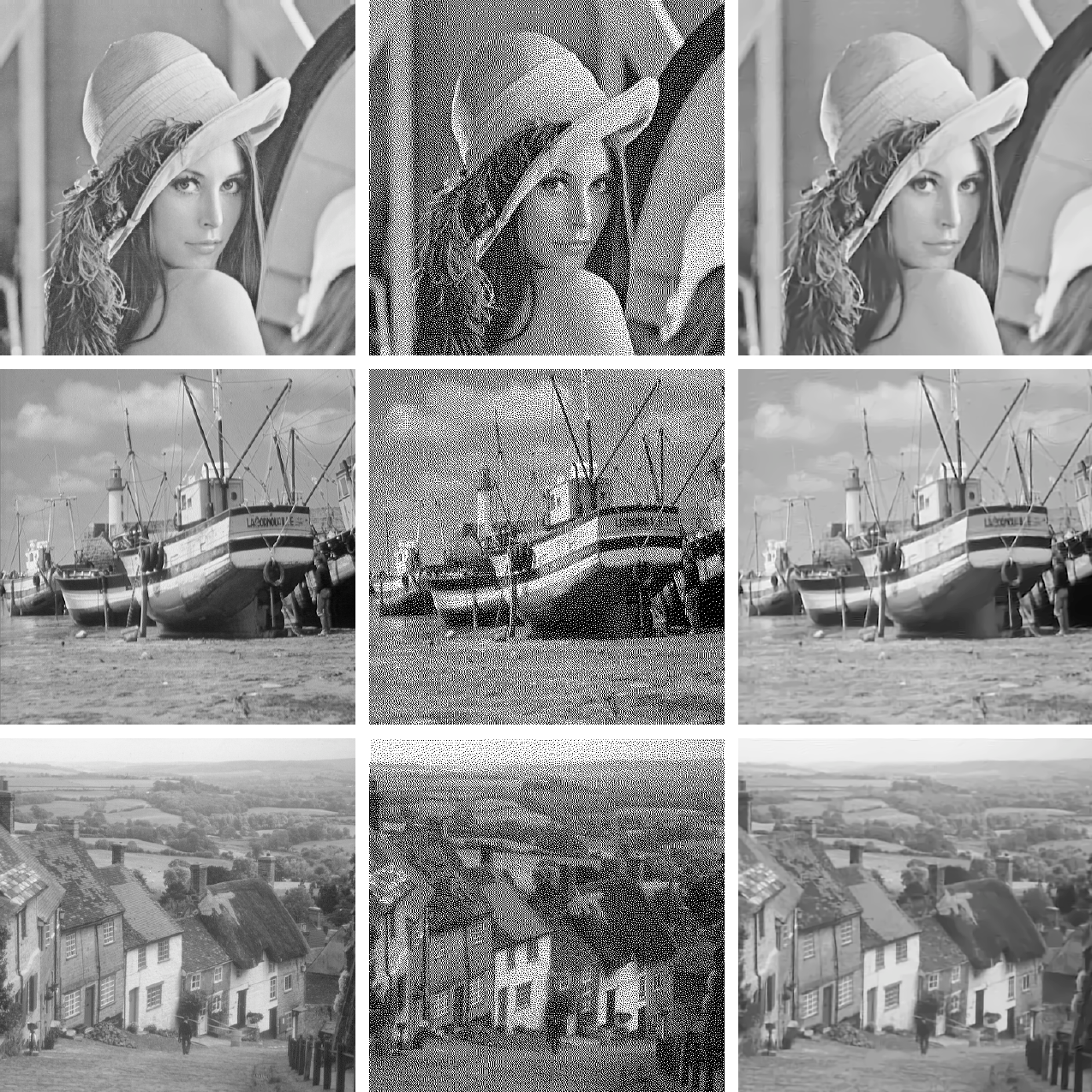} 
   \caption{From left to right: Original images, halftoned images, reconstructed images. Even though the halftoned images (center column) perceptually look relatively close to the original images (left column), they are binary. Reconstructed images (right column) are obtained by restoring the halftoned binary images. Best viewed by zooming on a computer screen.}
\label{fig:halftones}
\end{figure}

We now illustrate our method in a regression context by considering a
classical image processing task called ``inverse halftoning''.  With the
development of several \emph{binary} display technologies in the 70s
(including, for example, printers and PC screens), the problem of converting a grayscale
continuous-tone image into a binary one that looks perceptually similar to the
original one (``halftoning'') was posed to the image
processing community.  Examples of halftoned images obtained with the classical Floyd-Steinberg
algorithm~\citep{floyd} are presented in the second column of
Figure~\ref{fig:halftones}, with original images in the first column.
Restoring these binary images to continuous-tone ones~(``inverse halftoning'')
has become a classical problem~(see \cite{dabov} and references therein).

Unlike most image processing approaches that 
explicitly model the halftoning process, we formulate it as a
regression problem, \emph{without exploiting any prior on the task}. 
We use a database of $36$ images; $24$ are high-quality images from the Kodak PhotoCD
dataset\footnote{\url{http://r0k.us/graphics/kodak/}} and are used for training,
and~$12$ are classical images often used for evaluating image processing
algorithms;\footnote{The list of these
images can be found in \citep{mairal8}, where they are used for the problem of
image denoising.} the first four (\textsf{house}, \textsf{peppers},
\textsf{cameraman}, \textsf{lena}) are used for validation and the remaining
eight for testing.

We apply the Floyd-Steinberg algorithm
implemented in the LASIP Matlab toolbox\footnote{\url{http://www.cs.tut.fi/~lasip/}}
to the grayscale continuous-tone images in order to build our training/validation/testing set. 
We extract all pairs of patches from the original/halftoned images in
the training set, which provides us with a database of  approximately $9$
millions of patches.  We then use the ``signal regression'' formulation of
Eq.~(\ref{eq:regression}) to learn a dictionary~$\D$ and model
parameters~$\W$, by performing two passes of our algorithm over the $9$ million
training pairs.

At this point, we have learned how to restore a small patch from an image, but
not yet how to restore a full image. Following other patch-based approaches to
image restoration~\cite{elad}, we extract from a test image all patches
including overlaps, and
restore each patch \emph{independently} so that we get different estimates
for each pixel (one estimate for each patch the pixel belongs to).
These estimates are then averaged to reconstruct the full image,
 which has proven to give very good results in many image
restoration tasks~(see, e.g., \cite{elad,mairal8}). The final image is then post-processed using
the denoising algorithm of~\citet{mairal8} to remove possible artefacts.

We then measure how well it reconstructs the continuous-tone images from the halftoned ones in the test
set.  To reduce the number of hyperparameters, we have made
a few arbitrary choices:  We first use the Lasso
formulation for encoding the signals---that is, we set $\lambda_2=0$.
With millions of training samples, our model is unlikely to overfit and the
regularization parameter~$\nu$ is set to $0$ as well. The remaining free
parameters are the size~$m$ of
the patches, the size~$p$ of the dictionary and the regularization
parameter~$\lambda_1$.  These parameters are selected by minimizing the mean-squared error reconstruction on the validation set. 
We have tried patches of size $m=l \times l$, with $l
\in \{6,8,10,12,14,16\}$, dictionaries of sizes $p=100$, $250$ and $500$ 
, and determined $\lambda_1$ by first trying values on the logarithmic scale $10^i$, $i=-3,2$,
then refining this parameter on the scale $0.1,0.2,0.3,\ldots,1.0$.  The best
parameters found are $m=10 \times 10$, $p=500$ and $\lambda_1=0.6$.  Since the test
procedure is slightly different from the training one (the test includes an
averaging step to restore a full image whereas the training one does not), we have
refined the value of $\lambda_1$, trying different values
one an additive scale $\{0.4,0.45,\ldots,0.75,0.8\}$, and selected the value
$\lambda_1=0.55$, which has given the best result on the validation set.

\begin{figure}
   \centering
    \includegraphics[width=\linewidth]{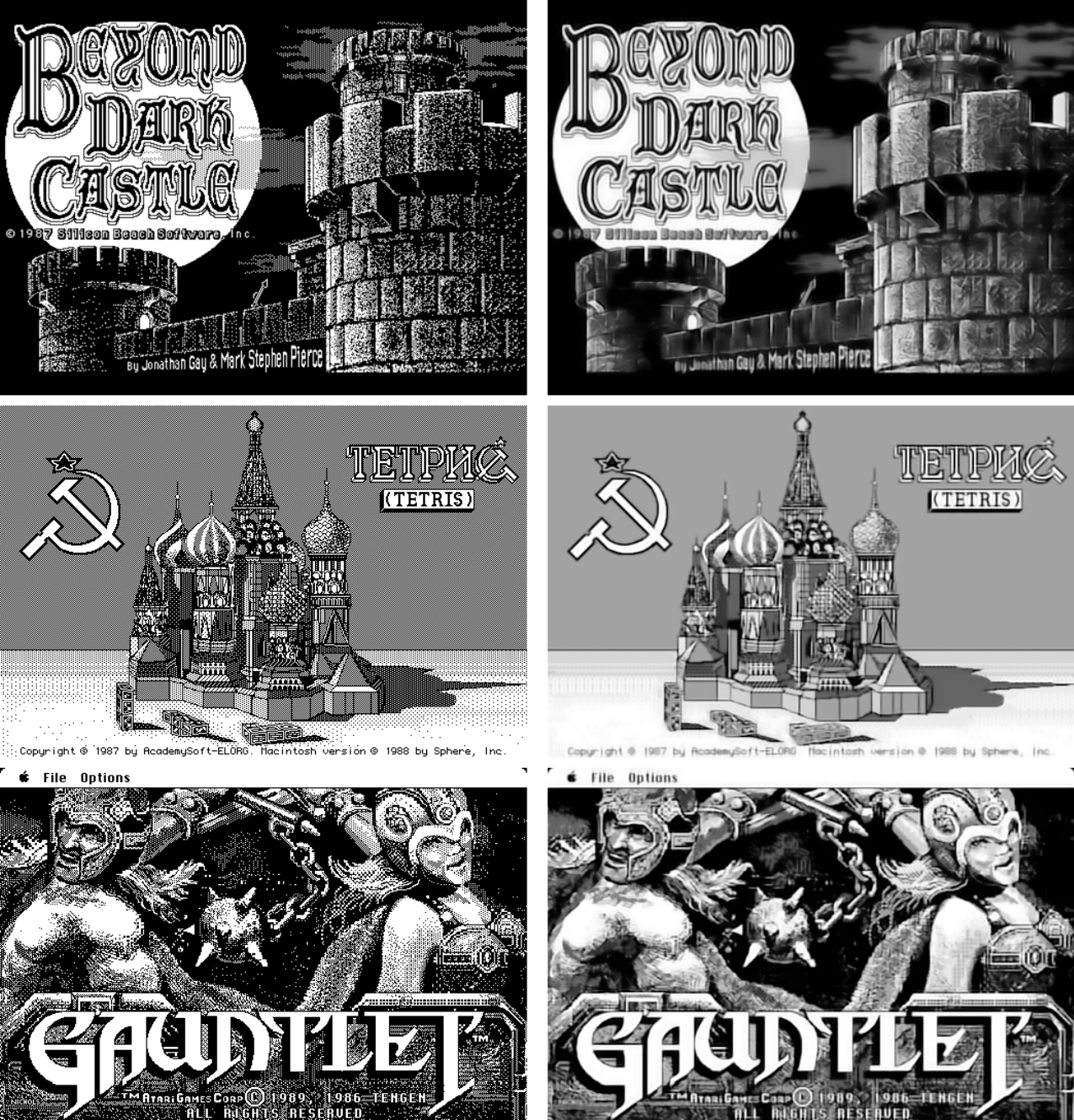} 
   \caption{Results on various binary images publicly available on the Internet. No ground truth is available for these images from old computer games, and the algorithm that has generated these images is unknown. Input images are on the left. Restored images are on the right. Best viewed by zooming on a computer screen.}
\label{fig:halftones2}
\end{figure}

Note that the largest dictionary has been chosen, and better results could potentially
be obtained using an even larger dictionary, but this would imply a higher
computational cost.  Examples of results are presented in
Figure~\ref{fig:halftones}. Halftoned images are binary but look perceptually
similar to the original image. Reconstructed images have very few artefacts and
most details are well preserved.  We report in
Table~\ref{table:halftones} a quantitative comparison between our approach and
various ones from the literature, including the state-of-the-art algorithm of
\cite{dabov}, which had until now the best results on this dataset.  Even
though our method does not explicitly model the transformation, it leads to
better results in terms of PSNR.\footnote{Denoting by \textrm{MSE} the
mean-squared-error for images whose intensities are between $0$ and $255$, the
\textrm{PSNR} is defined as $\textrm{PSNR}=10\log_{10}( 255^2 / \textrm{MSE} )$
and is measured in dB. A gain of $1$dB reduces the \textrm{MSE} by
approximately $20\%$.} We also present in Figure~\ref{fig:halftones2} the
results obtained by applying our algorithm to various binary images found on
the web, from which we do not know the ground truth, and which have not
necessarily been obtained with the Floyd-Steinberg algorithm. The results
are qualitatively rather good.

From this experiment, we conclude that our method is well suited to (at least, some) nonlinear regression problems on natural images, paving the way
to new applications of sparse image coding.
\begin{table*}
   \renewcommand{\arraystretch}{1.3}
\caption{Inverse halftoning experiments. Results are reported in PSNR (higher is better). SA-DCT refers to \citep{dabov}, LPA-ICI to \citep{foi}, FIHT2 to \citep{kite} and WInHD to \citep{neelamani}. Best results for each image are in bold. 
} \label{table:halftones}
\centering
\ifthenelse{\isundefined{\IEEEcompilation}}{
\vspace*{0.25cm}
}{ }
\begin{tabular}{*{13}{|c}|}
\hline
& \multicolumn{4}{c|}{Validation set} & \multicolumn{8}{c|}{Test set} \\
\hline
Image & 1 & 2 & 3 & 4& 5& 6& 7& 8& 9& 10& 11& 12 \\
\hline
FIHT2 & 30.8 & 25.3 & 25.8 & 31.4 & 24.5 & 28.6 & 29.5 & 28.2 & 29.3 & 26.0 & 25.2 & 24.7 \\
\hline
WInHD & 31.2 & 26.9 & 26.8 & 31.9 & 25.7 & 29.2 & 29.4 & 28.7 & 29.4 & 28.1 & 25.6 & 26.4 \\
\hline
 $\!$LPA-ICI$\!$ & 31.4 & 27.7 & 26.5 & 32.5 & 25.6 & 29.7 & 30.0 & 29.2 & 30.1 & 28.3 & 26.0 & 27.2 \\
\hline
 $\!$SA-DCT$\!$ & 32.4 & 28.6 & 27.8 & \textbf{33.0} & \textbf{27.0} & 30.1 & 30.2 & 29.8 & 30.3 & 28.5 & \textbf{26.2} & 27.6  \\
\hline
Ours & \textbf{33.0} & \textbf{29.6} & \textbf{28.1} & \textbf{33.0} & 26.6 & \textbf{30.2} & \textbf{30.5} & \textbf{29.9} & \textbf{30.4} & \textbf{29.0} & \textbf{26.2} & \textbf{28.0}  \\
\hline
\end{tabular}
\end{table*}

\subsection{Digital Art Authentification} \label{subsec:art}
Recognizing authentic paintings from imitations using statistical techniques
has been the topic of a few recent works~\citep{lyu,johnson,hugues}.  Classical
methods compare, for example, the kurtosis of wavelet coefficients between a
set of authentic paintings and imitations~\citep{lyu}, or involve more
sophisticated features~\citep{johnson}.  Recently, Hugues et al.~\cite{hugues}
have considered a dataset of $8$ authentic paintings from Pieter Bruegel the
Elder, and $5$ imitations.\footnote{The origin of these paintings is 
assessed by art historians.}  They have proposed to learn dictionaries for
sparse coding, and use the kurtosis of the sparse coefficients as
discriminative features.  We use their dataset, which they kindly provided to
us.\footnote{It would have been interesting to use the datasets used in \cite{lyu,johnson}, but they are not publicly available.}

The supervised dictionary learning approach we have presented is designed for
classifying relatively small signals, and should not be directly applicable to
the classification of large images, for which classical computer vision
approaches based on bags of words may be better
adapted~(see~\cite{yang,boureau} for such approaches).  However, we show that, for this particular
dataset, a simple voting scheme based on the classification of small image
patches with our method leads to good results.

The experiment we carry out consists of finding which painting is authentic,
and which one is fake, in a pair known to contain one of each.\footnote{This
task is of course considerably easier than classifying each painting as
authentic or fake. We do not claim to propose a method that readily applies to
forgery detection.} We proceed in a leave-one-out fashion, where we remove for
testing one authentic and one imitation paintings from the dataset, and learn
on the remaining ones. Since the dataset is small and does not have an official training/test set,
we measure a cross-validation score, testing all
possible pairs.  We consider $12 \times 12$ color image patches, without any
pre-processing, and classify each patch from the test images independently.
Then, we use a simple majority vote among the test patches to decide which
image is the authentic one in the pair test, and which one is the imitation.\footnote{Note that this experimental setting is different from
\cite{hugues}, where only authentic paintings are used for training (and not imitations). We therefore do not make quantitive comparison with this work.}

For each pair of authentic/imitation paintings, we build a dataset containing
$200\,000$ patches from the authentic images and $200\,000$ from the imitations.
We use the formulation from Eq.~(\ref{eq:linear}) for binary classification,
and arbitrarily choose dictionaries containing $p=100$ dictionary elements.
Since the training set is large, we set the parameter $\nu$ to $0$, 
also choose the Lasso formulation for decomposing the patches by setting $\lambda_2=0$, 
and cross-validate on the parameter $\lambda_1$, trying values on a
grid $\{10^{-4},10^{-3},\ldots,10^0\}$, and then refining the result on a grid
with a logarithmic scale of $2$.  We also compare Eq.~(\ref{eq:linear}) 
with the logistic regression loss and the basic formulation of
Eq.~(\ref{eq:unsup}) where $\D$ is learned unsupervised.

For classifying individual patches, the cross-validation score of the supervised formulation is
a classification rate of $54.04 \pm 2.26 \%$, which slightly improves upon the
``unsupervised'' one that achieves $51.94 \pm 1.92 \%$.  The task of
classifying independently small image patches is difficult since there is
significant overlap between the two classes.  On the other hand, finding the imitation
in a pair of (authentic,imitation) paintings with the voting scheme is easier
and the ``unsupervised formulation'' only fails for one pair, whereas the
supervised one has always given the right answer in our experiments.

\subsection{Compressed Sensing} \label{sec:expcs}
\begin{table*}[bt!]
   \caption{Compressed sensing experiment on small natural image patches.
   The mean squared error (MSE) measured on a test set is reported for
   each scenario with standard deviations, obtained by reproducing $5$
   times each experiment, randomizing the algorithm initializations and
   the sampling of the training images.  Each patch is normalized to have
   unit $\ell_2$ norm, and the mean squared reconstruction error is
   multiplied by $100$ for readability purposes. Here, $r$ is the number
   of rows of the matrix $\Z$. The different scenarios vary with the way
   $\D$ and $\Z$ are learned (fixed, unsupervised, supervised). See the main text for details.} \label{table:cs} 
   \renewcommand{\arraystretch}{1.3}
   \centering
\ifthenelse{\isundefined{\IEEEcompilation}}{
\small
\vspace*{0.25cm}
}{ }
\ifthenelse{\isundefined{\IEEEcompilation}}{
\begin{tabular}{|l|*{4}{|m{1.3cm}}|*{4}{|m{1.3cm}}|}
}{
  \begin{tabular}{|l|*{4}{|c}|*{4}{|c}|}
}
      \hline
      {\centering $\Z$} & \multicolumn{3}{c|}{\textsf{RANDOM}} & \textsf{SL1} & \multicolumn{3}{c|}{\textsf{PCA}} & \textsf{SL2} \tabularnewline
      \hline
      $\D$ & \textsf{DCT} & \textsf{UL} & \textsf{SL} & \textsf{SL}  & \textsf{DCT} & \textsf{UL} & \textsf{SL} & \textsf{SL}  \tabularnewline
      \hline
      $\!\! r=5 \!\!$ & $77.3 \pm 4.0$ &  $76.9 \pm 4.0$ &  $76.7 \pm 4.0$ &  $54.1 \pm 1.3$ &  $49.9 \pm 0.0$ &  $47.6 \pm 0.0$ &  $47.5 \pm 0.1$ &  $47.3 \pm 0.3$ \tabularnewline
      \hline
      $\!\! r=10 \!\!$ & $57.8 \pm 1.5$ &  $56.5 \pm 1.5$ &  $55.7 \pm 1.4$ &  $36.5 \pm 0.7$ &  $33.7 \pm 0.0$ &  $32.3 \pm 0.0$ &  $32.3 \pm 0.1$ &  $31.9 \pm 0.2$ \tabularnewline
      \hline
      $\!\! r=20 \!\!$ & $37.1 \pm 1.2$ &  $35.4 \pm 1.0$ &  $34.5 \pm 0.9$ &  $21.4 \pm 0.1$ &  $20.4 \pm 0.0$ &  $19.7 \pm 0.0$ &  $19.6 \pm 0.1$ &  $19.4 \pm 0.2$ \tabularnewline
      \hline
      $\!\! r=40 \!\!$ & $19.3 \pm 0.8$ &  $18.5 \pm 0.7$ &  $18.0 \pm 0.6$ &  $10.0 \pm 0.3$ &  $9.2 \pm 0.0$ &  $9.1 \pm 0.0$ &  $9.0 \pm 0.0$ &  $9.0 \pm 0.0$ \tabularnewline
      \hline
   \end{tabular}
\end{table*}

In this experiment, we apply our method to the problem of learning dictionaries
and projection matrices for compressed sensing.  As explained in
Section~\ref{subsec:cs}, our formulation and the conclusions of this section
hold for relatively small signals,  where the sensing matrix can be stored into
memory and learned.  Thus, we consider here small image patches of natural
images of size $m=10 \times 10$ pixels.  To build our training/validation/test
set, we have chosen the Pascal VOC 2006 database of natural
images~\citep{pascal2}: Images~$1$ to~$3000$ are used for training; images
$3001$ to $4000$ are used for validation, and the remaining $1304$ images are
kept for testing. Images are downsampled by a factor~$2$ so that the
JPEG compression artefacts present in this dataset become visually imperceptible,
thereby improving its quality for our experiment.

We compare different settings where the task is to reconstruct the
patches~$\y$ of size $m=10 \times 10$, from an observation~$\Z\x$ of size $r
\ll m$ (for instance $r=20$ linear measurements), where~$\Z$ in $\Real^{r \times m}$ is a sensing matrix.  For simplicity
reasons, we only consider here the noiseless case, where $\y = \x$.  At test
time, as explained in Section~(\ref{subsec:cs}), our estimate of~$\y$ is
$\W\alphab^\star(\Z\x,\D)$, where~$\D$ in~$\Real^{r \times p}$ is a dictionary
for representing~$\Z\x$, and~$\W$ in~$\Real^{m \times p}$ can be interpreted
here as a dictionary for representing~$\y$.
We evaluate several strategies for obtaining $(\Z,\D,\W)$:
\begin{itemize}
   \item We consider the case, which we call \textsf{RANDOM}, where the entries of~$\Z$ 
      are i.i.d. samples of the Gaussian distribution ${\mathcal N}(0,1/\sqrt{m})$.
      Since the purpose
      of~$\Z$ is to reduce the dimensionality of the input data, it is also natural
      to consider the case where $\Z$ is obtained by principal component analysis
      on natural image patches (\textsf{PCA}).
      Finally, we also learn~$\Z$ with the supervised learning
      formulation of Eq.~(\ref{eq:cs}), (\textsf{SL}), but consider the case where it is initialized randomly (\textsf{SL1})
      or by PCA (\textsf{SL2}).
   \item
   The matrix~$\D$ can either be fixed or learned.
      A typical setting would be to set $\D=\Z\D'$, where~$\D'$ is discrete-cosine-transform matrix
      (\textsf{DCT}), often used in signal processing applications~\citep{elad}. 
      It can also be learned with an unsupervised learning formulation (\textsf{UL}),
      or a supervised one (\textsf{SL}).
   \item $\W$ is always learned in a supervised way.
\end{itemize}

Since our training set is very large (several millions of
patches), we arbitrarily set the regularization parameters $\nu_1$ and $\nu_2$ to $0$. We
measure reconstruction errors with dictionaries of various levels of
overcompleteness by choosing a size $p$ in $\{100,200,400\}$.
The remaining free parameters are the regularization parameters $\lambda_1$ and~$\lambda_2$ for obtaining the coefficients $\alphab^\star$. We try the values
$\lambda_1 =10^i$, with $i$ in $\{-5,\ldots,0\}$.
Unlike what we have done in the experiments of Section~\ref{subsec:expartefacts},
it is absolutely necessary in this setting to use $\lambda_2 > 0$ (according to the theory), 
since using a zero value for this parameter has led to
instabilities and prevented our algorithm from converging. 
We have tried the values $\lambda_2 = 10^i \lambda_1$, with $i$ in
$\{-2,-1,0\}$.  Each learning procedure is performed by our algorithm in one
pass on $10$ millions of patches randomly extracted from our training images.
The pair of parameters $(\lambda_1,\lambda_2)$ that gives the lowest
reconstruction error on the validation set is selected, and we report in Table~\ref{table:cs} the
result obtained on a test set of $500\,000$ patches randomly extracted from the $1 304$ test images.
The conclusions of this compressed sensing experiment on natural image patches are the following:
\begin{itemize}
   \item {When $\Z$ is initialized as a Gaussian random matrix (case \textsf{RANDOM}), learning $\D$ and $\Z$ significantly improves the reconstruction error (case \textsf{SL1}).} A similar observation was made in~\citet{duarte}.
   \item {Results obtained with \textsf{PCA} are in general much better than those obtained with random projections}, which is consistent with the conclusions of~\citet{weiss}.
   \item {However, \textsf{PCA} does better than \textsf{SL1}. When \textsf{PCA} is used for initializing our supervised formulation, results can be slightly improved (case \textsf{SL2}).} This illustrates either the limits of the non-convex optimization procedure, or that \textsf{PCA} 
      is particularly well adapted to this problem.
   \item {Learned dictionaries (cases \textsf{UL} and \textsf{SL}) outperform classical \textsf{DCT} dictionaries.}  
\end{itemize}
\section{Conclusion} \label{sec:ccl}
We have presented in this paper a general formulation for learning sparse
data representations tuned to specific tasks.  Unlike classical approaches
that learn a dictionary adapted to the reconstruction of the input data,
our method learns features in a supervised way.  We have shown that this
approach is effective for solving classification and regression tasks in a
large-scale setting, allowing the use of millions of training samples, and
is able of exploiting successfully unlabeled data, when only a few labeled
samples are available.  Experiments on handwritten digits classification,
non-linear inverse image mapping, digital art authentification, and
compressed sensing have shown that our method leads to state-of-the-art
results for several real problems.  Future work will include adapting our
method to various image processing problems such as image deblurring and
image super-resolution, and other inverse problems.
\appendix[Proofs and Lemmas]
Before giving the proof of Proposition~\ref{prop:regf}, we present
two general results on the elastic net formulation of \citet{zou}.
\UpperSpace
\begin{lemma}[Optimality conditions of the elastic net] \label{lemma:elas} ~\newline
   The vector~$\alphab^\star$ is a solution of Eq.~(\ref{eq:elas}) if and only if for all
   $j$ in $\{1,\ldots,p\}$, \begin{equation}
   \begin{split}
       \d_j^\top(\x-\D\alphab^\star) - \lambda_2 \alphab^\star[j] &=\lambda_1
       \sign(\alphab^\star[j]) ~~\text{if}~~ \alphab^\star[j] \neq 0, \\
       |\d_j^\top(\x-\D\alphab^\star) - \lambda_2 \alphab^\star[j]|
       &\leq\lambda_1 ~~\text{otherwise}.
    \end{split} \label{eq:opt}
 \end{equation}
 Denoting by $\Lambda \defin \{ j \in \{1,\ldots,p\} \st \alphab^\star[j] \neq
 0 \}$ the active set, we also have  
 \begin{equation}
    \alphab_\Lambda^\star =
    (\D_\Lambda^\top\D_\Lambda+\lambda_2\I)^{-1}(\D_\Lambda^\top\x-\lambda_1
    \s_\Lambda), \label{eq:closed}
 \end{equation}
 where $\s_\Lambda$ in $\{-1; +1\}^{|\Lambda|}$ carries the signs of
 $\alphab^\star_\Lambda$.
\end{lemma}
\LowerSpace
\begin{IEEEproof}
   Equation~(\ref{eq:opt}) can be obtained by considering subgradient optimality
   conditions as done in \citet{fuchs} for the case $\lambda_2=0$.  
   These can be written as
   \begin{displaymath}
      0 \in \{ -\D^\top(\x-\D\alphab^\star) + \lambda_2 \alphab^\star +
      \lambda_1 \p  :  \p  \in \partial \|\alphab^\star\|_1 \},
   \end{displaymath}
   where $\partial\|\alphab^\star\|_1$ denotes the subdifferential of the $\ell_1$
   norm evaluated at $\alphab^\star$.  A classical result~(see~\cite{borwein},
   page 238) is that the subgradients $\p$ of this subdifferential are
   characterized by the fact that for all~$j$ in
   $\{1,\ldots,p\}$, $\p[j]=\sign(\alphab^\star[j])$ if $\alphab^\star[j] \neq
   0$, and $|\p[j]| \leq 1$ otherwise.  This gives directly Eq.~(\ref{eq:opt}).
   The equalities in Eq.~(\ref{eq:opt}) define a linear system whose solution
   is Eq.~(\ref{eq:closed}).
\end{IEEEproof}
The next proposition exploits these optimality conditions to characterize
the regularity of $\alphab^\star$.
\UpperSpace
\begin{proposition}[Regularity of the elastic net solution] \label{prop:elasreg} ~\\
   Assume $\lambda_2 > 0$ and ({\bf A}). Then,
   \begin{enumerate}
      \item The function $\alphab^\star$ is uniformly Lipschitz on $\XX \times \DD$.
      \item Let $\D$ be in $\DD$, $\varepsilon$ be a positive scalar and $\s$ be a vector in $\{-1,0,+1\}^p$, and 
         define $K_\s(\D,\varepsilon) \subseteq \XX$ as the set of vectors $\x$ satisfying for all $j$ in $\{1,\ldots,p\}$,
         \begin{equation}
            \left\{ 
            \begin{array}{lcr}
               |\d_j^\top(\x-\D\alphab^\star) - \lambda_2 \alphab^\star[j]| \leq \lambda_1-\varepsilon &\text{if}& \s[j] = 0, \\
               \s[j]\alphab^\star[j] \geq \varepsilon & \text{if}& \s[j] \neq 0. \\
            \end{array}
            \right. \label{eq:Knub}
         \end{equation}
         where $\alphab^\star$ is shorthand for $\alphab^\star(\x,\D)$.

         Then, there exists $\kappa > 0$ independent of $\s$, $\D$ and $\varepsilon$ so
         that for all $\x$ in $K_\s(\D,\varepsilon)$,
         the function~$\alphab^\star$ is twice continuously differentiable on
         $B_{\kappa\varepsilon}(\x) \times B_{\kappa\varepsilon}(\D)$,  where
         $B_{\kappa\varepsilon}(\x)$ and $B_{\kappa\varepsilon}(\D)$ denote the
         open balls of radius $\kappa\varepsilon$ respectively centered on $\x$
         and $\D$.
   \end{enumerate}
\end{proposition}
\LowerSpace
\begin{IEEEproof}
   The first point is proven in \cite{mairal7}. The proof uses
   the strong convexity induced by the elastic-net term, when $\lambda_2 >
   0$, and the compactness of $\XX$ from Assumption~({\bf A}).

   For the second point, we study the differentiability of $\alphab^\star$ on 
   sets that satisfy conditions which are more restrictive than the optimality
   conditions of Eq.~(\ref{eq:opt}).  Concretely, let~$\D$ be in~$\DD$,
   $\varepsilon > 0$ and~$\s$ be in $\{-1,0,+1\}^p$.  The set
   $K_\s(\D,\varepsilon)$ characterizes the vectors~$\x$ so that
   $\alphab^\star(\x,\D)$ has the same signs as~$\s$ (and same set of zero
   coefficients), and $\alphab^\star(\x,\D)$ satisfies the conditions of
   Eq.~(\ref{eq:opt}), but with two additional constraints: (i) The magnitude
   of the non-zero coefficients in $\alphab^\star$ should be greater than~$\varepsilon$. (ii) The inequalities in Eq.~(\ref{eq:opt}) should be strict
   with a margin $\varepsilon$.  The reason for imposing these assumptions is to
   restrict ourselves to points~$\x$ in~$\XX$ that have a stable active set---that is,
   the set of non-zero coefficients~$\Lambda$ of~$\alphab^\star$ should not
   change for small perturbations of $(\x,\D)$, when $\x$ is in
   $K_\s(\D,\varepsilon)$.
   
   Proving that there exists a constant $\kappa > 0$ satisfying the second
   point is then easy (if a bit technical): Let us assume that
   $K_\s(\D,\varepsilon)$ is not empty (the case when it is empty is
   trivial).
   Since~$\alphab^\star$ is
   uniformly Lipschitz with respect to $(\x,\D)$, so are the quantities
   $\d_j^\top(\x-\D\alphab^\star)-\lambda_2\alphab^\star[j]$ and
   $\s[j]\alphab^\star[j]$, for all $j$ in $\{1,\ldots,p\}$.
   Thus, there exists $\kappa > 0$ independent of $\x$ and
   $\D$ such that for all $(\x',\D')$ satisfying $\|\x-\x'\|_2 \leq
   \kappa\varepsilon$ and $\|\D-\D'\|_F \leq \kappa\varepsilon$, we have 
   for all $j$ in $\{1,\ldots,p\}$,
   \begin{displaymath}
 \left\{ 
      \begin{array}{lcr}
         |\d_j^{\top\prime}(\x'-\D'\alphab^{\star\prime}) - \lambda_2 \alphab^{\star\prime}[j]| \leq \lambda_1-\frac{\varepsilon}{2} &\text{if}& \s[j] = 0, \\
         \s[j]\alphab^{\star\prime}[j] \geq \frac{\varepsilon}{2} & \text{if}& \s[j] \neq 0. \\
      \end{array}
      \right.
   \end{displaymath}
   where $\alphab^{\star\prime}$ is short-hand for $\alphab^\star(\x',\D')$,
   and $\x'$ is therefore in $K_{\s}(\D',\varepsilon/ 2)$.
   It is then easy to show that the active set $\Lambda$ of $\alphab^\star$
   and the signs of $\alphab^\star$ are stable on $B_{\kappa\varepsilon}(\x)
   \times B_{\kappa\varepsilon}(\D)$, and that $\alphab_\Lambda^\star$ is given
   by the closed form of Eq.~(\ref{eq:closed}).  $\alphab^\star$ is therefore
   twice differentiable on $B_{\kappa\varepsilon}(\x)
   \times B_{\kappa\varepsilon}(\D)$.
\end{IEEEproof}
With this proposition in hand, we can now present the proof of Proposition~\ref{prop:regf}: 
\begin{IEEEproof}
   The differentiability of $f$ with respect to $\W$ is easy using
   only the compactness of~$\YY$ and~$\XX$ and the fact that $\l_s$ is twice differentiable.
   We will therefore focus on showing that $f$ is differentiable with
   respect to $\D$, which is more difficult since $\alphab^\star$ is \emph{not}
   differentiable everywhere.

   Given a small perturbation $\EE$ in $\Real^{m \times p}$ of $\D$, we 
   compute
   \begin{multline}
      f(\D+\EE,\W)-f(\D,\W) = \\ \E_{\y,\x}\Big[\nabla_\alphab \l_s^\top \big(\alphab^\star(\x,\D+\EE)-\alphab^\star(\x,\D)\big) \Big] + O(\|\EE\|_\FFF^2), \label{eq:proof}
   \end{multline}
   where $\nabla_\alphab \l_s$ is short for $\nabla_\alphab \l_s(\y,\W,\alphab^\star)$, and
   the term $O(\|\EE\|_\FFF^2)$ comes from the fact that $\alphab^\star$ is
   uniformly Lipschitz and $\XX \times \DD$ is compact.

   Let now choose $\W$ in $\WW$ and $\D$ in $\DD$.  We have characterized in
   Lemma~\ref{prop:elasreg} the differentiability of~$\alphab^\star$ on some
   subsets of $\XX \times \DD$. We consider the set 
    \begin{displaymath}
       K(\D,\varepsilon) \defin 
       \bigcup_{\s \in \{-1,0,1\}^p} K_\s(\D,\varepsilon),
    \end{displaymath}
    and denoting by $\PPP$ our probability measure, it is easy to show with a few calculations that
    $\PPP(\XX \setminus K(\D,\varepsilon)) = O(\varepsilon)$.
    Using the constant $\kappa$ defined in Lemma~\ref{prop:elasreg}, 
    we obtain that $\PPP(\XX \setminus K(\D,\|\EE\|_\FFF/\kappa)) =
    O(\|\EE\|_\FFF)$. 
    Since $\nabla_\alphab \l_s(\y,\W,\alphab^\star)^\top \big(\alphab^\star(\x,\D+\EE)-\alphab^\star(\x,\D)\big)=O(\|\EE\|_\FFF)$, the set  $\XX \setminus K(\D,\|\EE\|_\FFF/\kappa)$ can be neglected (in the formal sense) when integrating with respect to~$\x$ in the expectation of Eq.~(\ref{eq:proof}), and it is possible to show that
    \begin{displaymath}
       f(\D+\EE,\W)-f(\D,\W) = \trace\big( \EE^\top g(\D,\W)\big) + O(\|\EE\|_F^2),
    \end{displaymath}
    where $g$ has the form given by Eq.~(\ref{eq:gradient}). This shows that $f$ is differentiable with respect to $\D$, and its gradient $\nabla_\D f$ is $g$.
\end{IEEEproof}

\section*{Acknowledgments}
This paper was supported in part by ANR under grant MGA ANR-07-BLAN-0311 and the European Research Council (SIERRA and VideoWorld Project).
Julien Mairal is now supported by the NSF grant SES-0835531 and NSF award CCF-0939370.
The authors would like to thank J.M. Hugues and D.J. Graham and D.N. Rockmore for providing us with the Bruegel dataset used
in Section~\ref{subsec:art}, and Y-Lan Boureau and Marc'Aurelio Ranzato for providing their experimental setting for the digit recognition task.

\ifCLASSOPTIONcaptionsoff
\newpage
\fi

\bibliographystyle{IEEEtran}
\bibliography{mairal_pami}

\begin{IEEEbiography}[{\includegraphics[width=1in,height=1.25in,clip,keepaspectratio]{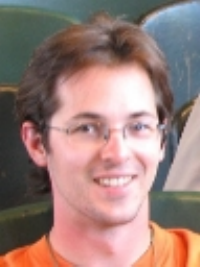}}]{Julien Mairal}
received the graduate degree from Ecole Polytechnique, Palaiseau, France in 2005 
and the PhD degree from the Ecole Normale Sup\'erieure, Cachan, France in 2010 under the supervision of Jean Ponce and Francis Bach in the INRIA Willow project-team in Paris. He recently joined the department of statistics of the University of California, Berkeley as a post-doctoral researcher. His research interests include machine learning, computer vision and image/signal processing. 
\end{IEEEbiography}

\begin{IEEEbiography}[{\includegraphics[width=1in,height=1.25in,clip,keepaspectratio]{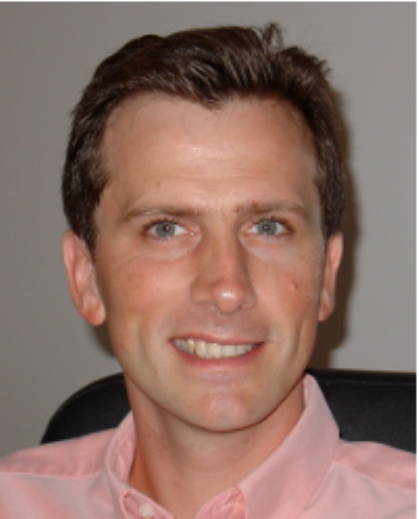}}]{Francis Bach}
is the leading researcher of the Sierra INRIA project-team, in the Computer Science Department of the Ecole Normale Sup\'erieure, Paris, France. He graduated from the Ecole Polytechnique, Palaiseau, France, in 1997, and earned his PhD in 2005 from the Computer Science division at the University of California, Berkeley. His research interests include machine learning, statistics, optimization, graphical models, kernel methods, and statistical signal processing.
\end{IEEEbiography}

\begin{IEEEbiography}[{\includegraphics[width=1in,height=1.25in,clip,keepaspectratio]{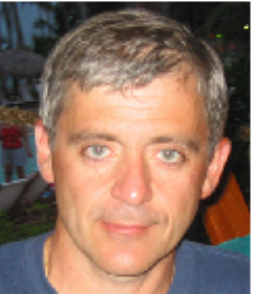}}]{Jean Ponce}
is a computer science professor at Ecole Normale Sup\'erieure (ENS) in Paris, France, where he heads the ENS/INRIA/CNRS Project-team WILLOW. Before joining ENS, he spent most of his career in the US, with positions at MIT, Stanford, and the University of Illinois at Urbana-Champaign, where he was a full professor until 2005. Jean Ponce is the author of over 120 technical publications in computer vision and robotics, including the textbook ``Computer Vision: A Modern Approach''. He is an IEEE Fellow, served as editor-in-chief for the International Journal of Computer Vision from 2003 to 2008, and chaired the IEEE Conference on Computer Vision and Pattern Recognition in 1997 and 2000, and the European Conference on Computer Vision in 2008.
\end{IEEEbiography}

\end{document}